\documentclass[runningheads]{llncs}

\usepackage[T1]{fontenc}
\usepackage{graphicx,verbatim}
\usepackage{amsmath}
\usepackage{amsfonts}
\usepackage{mathtools}
\usepackage{booktabs}
\usepackage{multicol}
\usepackage{multirow}
\usepackage{array}
\usepackage{tabularx}
\usepackage[hidelinks]{hyperref}

\begin{document}
\title{Ordinal Diffusion Models for \\ Color Fundus Images}

\author{
Gustav Schmidt \and
Philipp Berens \and
Sarah Müller
}

\authorrunning{G. Schmidt et al.}
\institute{
    Hertie Institute for AI in Brain Health, University of Tübingen, Germany
\email{sar.mueller@uni-tuebingen.de}}
  
\maketitle 

\begin{abstract}
Generative image models such as diffusion models can improve performance on clinically relevant tasks by offering deep learning models supplementary training data. However, most conditional diffusion models treat disease stages as independent classes, ignoring the continuous nature of disease progression. This mismatch is problematic in medical imaging because continuous pathological processes are typically only observed through coarse, discrete but ordered labels as in ophthalmology for diabetic retinopathy (DR). We propose an ordinal latent diffusion model for generating color fundus images that explicitly incorporates the ordered structure of DR severity into the generation process. Instead of categorical conditioning, we used a scalar disease representation, enabling a smooth transition between adjacent stages. We evaluated our approach using visual realism metrics and classification-based clinical consistency analysis on the EyePACS dataset. Compared to a standard conditional diffusion model, our model reduced the Fréchet inception distance for four of the five DR stages and increased the quadratic weighted $\kappa$ from 0.79 to 0.87. Furthermore, interpolation experiments showed that the model captured a continuous spectrum of disease progression learned from ordered, coarse class labels\footnote{\url{https://github.com/berenslab/OrdinalDiffusionModels}}.
\end{abstract}

\section{Introduction}

\begin{figure}[t!]
    \centering
    \includegraphics[width=\textwidth]{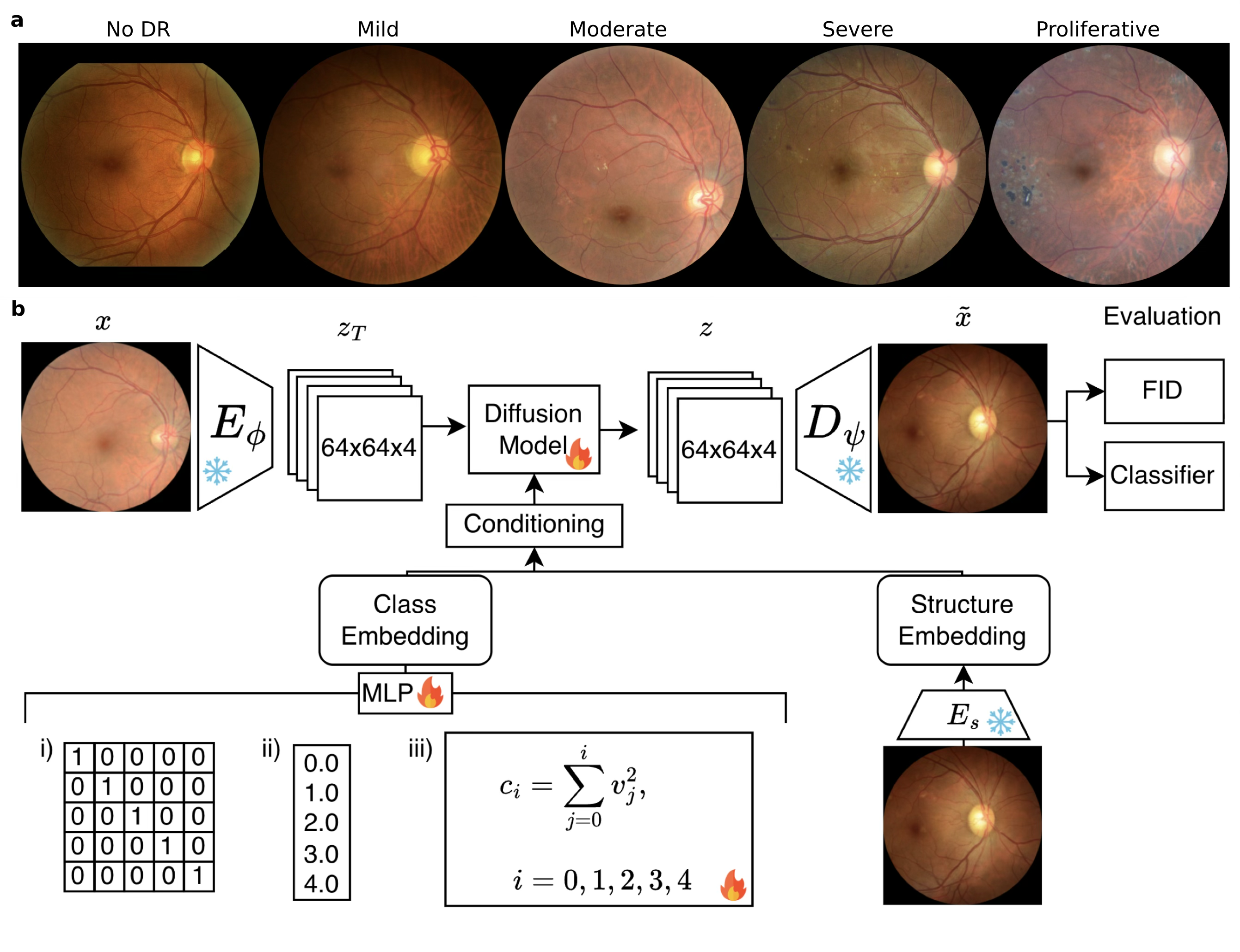}
    \caption{Overview figure of the proposed approach. Our diffusion model generates fundus images with different disease severity (\textbf{b}) by conditioning on ordinal embeddings (\textbf{a}). Symbols indicate jointly trained (flame) and pretrained frozen (snowflake) components.}
    \label{fig:ov}
\end{figure}

Diabetic retinopathy (DR) is a leading cause of preventable vision loss worldwide, and it is routinely monitored using retinal color fundus photography (CFP). These images allow a direct, noninvasive view of the microvasculature and neural tissue; this is why fundus imaging is important in large-scale screening programs~\cite{London2012}. DR progression is continuous, with small anatomical and functional changes accumulating over time. However, clinicians generally follow a severity scale consisting of five ordinal stages: No DR (stage 0), mild non-proliferative DR (stage 1), moderate NPDR (stage 2), severe NPDR (stage 3) and proliferative DR (stage 4) (Fig.~\ref{fig:ov}\textbf{a}). 
Although deep learning models have achieved excellent performance in DR detection in general~\cite{nielsen2019deep}, their development and application is restricted by access to sufficiently large diverse datasets, in particular with sufficient examples of late disease stages or from persons with underrepresented ethnicity~\cite{nakayama2023artificial}. Generative models may offer a solution by synthesizing realistic fundus images, which have been suggested to improve the performance of deep learning models in medical imaging providing additional data for training on underrepresented classes~\cite{khosravi2024synthetically,akrout2023diffusion,cheng2023robust}. While many attempts at generative modeling for CFPs have mostly resorted to generative adversarial networks (GANs)~\cite{hou2023high,muller2025disentangling,lang2021explaining}, diffusion models have demonstrated superior stability, diversity and scalability~\cite{Dhariwal,rombach,ilanchezian2025development}. For example, \cite{ilanchezian2025development} used classifier-guidance for training a diffusion model for CFP generation, requiring robust classifiers to be trained in the first place. However, the ordinal disease labels could already provide sufficient information for generating clinically meaningful, realistic fundus images along the DR disease spectrum. 
In this work, we present an ordinally conditioned latent diffusion model for generating CFP images that explicitly models the continuous progression of DR (Fig.~\ref{fig:ov}\textbf{b}). First, we explicitly encode the ordered structure of DR severity into the diffusion process, enabling smooth transitions between DR stages and modeling disease progression as a continuum rather than discrete categories. Second, we propose a dual-conditioning strategy that separates anatomical retinal structure from disease-specific pathology via a learned structural embedding, improving visual realism, disease consistency, and controllable interpolation across severity levels. Finally, we evaluate the model using both standard image quality metrics and classifier-based disease consistency analysis to verify that generated images preserve clinically meaningful severity ordering.

\section{Related Work}
Ordinal learning~\cite{shin2022moving,wang2023ord2seq} has received comparably little attention in medical imaging. It has been studied in a discriminative setting to model ordered labels for disease severity grading~\cite{liu2018ordinal,yu2024clip,cheng2023robust}. These methods reduce clinically implausible errors by explicitly introducing ordinal constraints, leading to increased performance and interpretability~\cite{Wang,schmidt2025learning}. Recently, a few others have developed generative models for medical data exploiting ordinal constraints~\cite{Kurt2025,Takezaki}. For example, \cite{Takezaki} suggest to employ geometric constraints on the difference of the noisy images to enforce ordinal ranks in latent space. Based on their evaluation, this approach did not consistently yield better performance in terms of realism than a simple diffusion model. We were unable to reproduce their findings because no code was provided, despite our best efforts to reimplement their approach. Similarly, \cite{Kurt2025} learn a high-dimensional look-up table to encode disease stages for endoscopic images based on a Stable Diffusion model. However, since no constraints are enforced on the look-up table entries, it is not clear how this approach can be considered ordinal at all. Thus, to our knowledge, our model is the first reproducible work that explicitly encodes ordinal geometric constraints into the conditioning representation of a latent diffusion model for disease staging.

\section{Methods}
\subsection{Ordinal-Aware Latent Diffusion Model}
\paragraph{Latent Autoencoding and Diffusion.}
We followed the standard latent diffusion framework~\cite{rombach}. An image $x \in \mathbb{R}^{H \times W \times 3}$ is mapped to a lower-dimensional latent using a Variational Autoencoder (VAE) with $f$ as the downsampling factor:
\[
    E_\phi : \mathbb{R}^{H \times W \times 3} \to \mathbb{R}^{h \times w \times c}, 
    \quad h = H/f, \; w = W/f.
\]
The encoder parametrizes a diagonal Gaussian posterior in latent space $E_\phi(x)=(\mu_\phi(x), \sigma_\phi(x)), \quad q_\phi(z \mid x) = \mathcal{N}\big(z; \mu_\phi(x), \sigma_\phi^2(x) I\big)$, with $z_0 \sim q_\phi(z \mid x) = \mu_\phi(x) + \sigma_\phi(x) \odot \epsilon, \quad \epsilon \sim \mathcal{N}(0,I)$, such that the diffusion model operates on stochastic latents. The VAE, consisting of encoder $E_\phi$ and decoder $D_\psi$, is trained via
\[
    \mathcal{L}_{\mathrm{AE}} =
    \mathbb{E}_{x}\Big[
    \|x-D_\psi(z_0)\|_1
    + \lambda_{\mathrm{perc}} \mathcal{L}_{\mathrm{perc}}(x,D_\psi(z_0))
    + \lambda_{\mathrm{KL}} D_{\mathrm{KL}}(q_\phi(z\mid x)\| \mathcal{N}(0,I))
    \Big],
\]
where $\mathcal{L}_{\mathrm{perc}}$ is a perceptual loss~\cite{johnson2016perceptual} and the KL term regularizes the latent space toward a standard normal. Forward diffusion is applied to $z_0$ over $t = 1,\dots,T$:
\[
    z_t = \sqrt{\bar{\alpha}_t}\, z_0 + \sqrt{1-\bar{\alpha}_t}\, \epsilon, 
    \quad \epsilon \sim \mathcal{N}(0,I), \quad
\]
with $\alpha_t=1-\beta_t$ and $\bar{\alpha}_t=\textstyle\prod_{k=1}^t \alpha_k$ and $\{\beta_t\}_{t=1}^T\subset(0,1)$ as the noise schedule.

\paragraph{Conditioning the Diffusion Model on Disease Stage and CFP Structure.}
The denoiser $\epsilon_\theta$ was conditioned on the timestep $t$, a representation of the disease grade $c$, and, optionally, the image structure $s$, and trained via
\[
    \mathcal{L}(\theta)=
    \mathbb{E}_{x_0,\epsilon,t}
    \left[
    \|\epsilon-\epsilon_\theta(z_t,t,c,s)\|_2^2
    \right],
    \quad t\sim\mathcal{U}\{1,\dots,T\},
\]
with random conditioning dropout for classifier-free guidance~\cite{ho2022classifier}. As a baseline, we encoded the disease grade $c$ with one-hot encoding; ordinal-aware representations are introduced below. To represent the structure of the image $s$ independently of pathology, we followed \cite{cheng2023robust} trained a structural encoder, $E_s:\mathbb{R}^{H \times W \times 3}\to\mathbb{R}^d, \quad s=E_s(x)$, via contrastive learning, implemented as a ResNet-50~\cite{he2016deep} trained with NT-Xent loss~\cite{chenSimpleFrameworkContrastive2020}. 
During sampling, we combined unconditional, disease-conditioned, and structure-conditioned predictions as
\[
    \epsilon_\theta(z_t,t,c,s)
    =\epsilon_\theta(z_t,t)
    +w_c\!\left(\epsilon_\theta(z_t,t,c)-\epsilon_\theta(z_t,t)\right)
    +w_s\!\left(\epsilon_\theta(z_t,t,s)-\epsilon_\theta(z_t,t)\right),
\]
where $w_c$ and $w_s$ control disease and structural guidance strength.

\paragraph{Ordinal Disease Stage Conditioning.}
In addition to the standard categorical one-hot encoding for disease grade conditioning, we implemented two strategies that explicitly capture disease stage ordering:
\begin{enumerate}
    \item \textbf{Equidistant margins:} disease stages are embedded on a 1D ordinal axis with equal spacing,
    $c_i = i, \quad i \in \{0,\dots,K-1\}$.
    \item \textbf{Learned margins:} relative spacings are learned via $c_i = \sum_{j=0}^{i} v_j^2$, $v \in \mathbb{R}^{K}$, where squaring enforces positive increments and guarantees monotonicity.
\end{enumerate}
In all cases, $c$ was passed through a two-layer fully connected MLP to produce a 1,024-dimensional embedding before being injected into the denoising network.

\subsection{Evaluation of Image Realism and Disease Consistency}
Visual realism was measured by the Fréchet inception distance (FID) score~\cite{Heusel2017-ps,parmarAliasedResizingSurprising2022}. Because FID is highly sensitive to small sample sizes, class-specific FID comparisons using per-class reference pools would be unreliable. We therefore computed FID using a fixed reference pool of 10,000 randomly sampled real images from the full dataset, compared against 10,000 generated images from each DR grade. Disease consistency was assessed using a pretrained DR classifier (ResNet-50 with the CORAL ordinal regression loss~\cite{caoRankConsistentOrdinal2020}) on the generated images. We computed the quadratic weighted $\kappa$ (QWK) between the predicted grades and the target stages used for generation. QWK penalizes disagreements according to their distance in the ordered disease-stage space.

\subsection{Dataset and Preprocessing}
We used the EyePACS dataset~\cite{Cuadros2009} (accessed 2022), larger than the public Kaggle subset. After quality filtering~\cite{Gervelmeyer2025}, 127,144 fundus images from 35,947 patients remained, with the following class distribution: Grade 0: 100,584; Grade 1: 9,948; Grade 2: 14,823; Grade 3: 960; Grade 4: 829. Images were center-cropped and resized to $256 \times 256$ pixels. We used task-specific strong augmentations (random crops, rotations, color jitter) for the DR classifier and the structural encoder $E_s$, while anatomically preserving augmentations (small rotations, horizontal flips, mild color jitter) were used for diffusion training. Data were split by patient identifier to avoid leakage. For latent diffusion and autoencoder training, we used a 90/10 train/validation split. To train the DR classifier, data were split 80/10/10 into train/validation/test sets. Weighted random sampling mitigated class imbalance. The final DR classifier, used to assess disease consistency, achieved a QWK of $0.63$ on the test set.

\subsection{Training Details}
All models were implemented in PyTorch~\cite{Falcon_PyTorch_Lightning_2019} (see code repository for full configuration files) and trained on NVIDIA A100 GPUs. 
The autoencoder was trained for 100 epochs using Adam ($\text{lr}=10^{-5}$, batch size $16$, $\lambda_{\mathrm{perc}}=0.25$), and the structural encoder for 100 epochs using Adam ($\text{lr}=10^{-4}$, batch size $256$). Diffusion was performed in a $64 \times 64 \times 4$ ($f=4$) latent space with a linear noise schedule over $T=1,000$ steps ($\beta_{\text{start}}=10^{-4}$, $\beta_{\text{end}}=0.02$). The denoiser was optimized using AdamW ($\text{lr}=10^{-5}$, batch size $16$). Sampling employed DDIM~\cite{songDenoisingDiffusionImplicit2022} with 100 inference steps and classifier-free guidance for both conditioning signals. Guidance weights were set empirically: $(w_c,w_s)=(4.0,0.0)$ for unconditional structural generation and $(7.0,1.5)$ for structure-preserving synthesis.

\section{Results}
\subsection{Ordinal Embedding Improved Realism and Disease Consistency}

We evaluated six model variants: a baseline model with categorical one-hot encoding for disease stages and two ordinally conditioned models with equidistant disease stages or learned distance between disease stages, each trained with and without structural conditioning (Table~\ref{tab:res2}). We found that for models \textit{without structural conditioning} both ordinal variants showed more realism as measured by the FID score than the baseline model for all DR grades except severe DR, where the model with equidistant margins was only slightly better than the baseline and the model with learned margins was slightly worse. In addition, the model with equidistant margins resulted in a QWK of 0.85 compared to 0.79 for the baseline model, indicating that learned disease features were more consistent with those of real images, while learned margins performed similar to the baseline. Including \textit{structural embeddings} into the model, we found again that the model with equidistant margins created more realistic images overall which were also more in line with disease features, except for proliferative DR, where apparently structural features interfered with disease features. The learned margin model did show much improved disease consistency, but the realism of the pictures was worse than without structural constraints. In the following analysis, we only considered model versions with structural conditioning.

\begin{table}[htbp]
    \centering
    \caption{Quantitative evaluation: class-wise FID and QWK for all model variants.}
    \label{tab:res2}
    \renewcommand{\arraystretch}{1}
        \begin{tabularx}{\linewidth}{l *{5}{>{\centering\arraybackslash}X} c}
        \toprule
        \multirow{2}{*}{\textbf{Setting}} 
        & \multicolumn{5}{c}{\textbf{FID $\downarrow$}} 
        & \multirow{2}{*}{\textbf{QWK $\uparrow$}} \\
        \cmidrule(lr){2-6}
        & No DR & Mild & Moderate & Severe & Prolif. & \\
        \midrule
        \textit{Baseline} & & & & & & \\
        \quad w/o structural cond. 
        & 22 & 23 & 19 & 16 & 63 & 0.79 \\
        \quad w/ structural cond.  
        & 23 & 22 & 23 & 18 & 22 & 0.82 \\
        \midrule
        \textit{Equidistant Margins} & & & & & & \\
        \quad w/o structural cond. 
        & 11 & 13 & 13 & 15 & 29 & 0.85 \\
        \quad w/ structural cond.  
        & 12 & 12 & 12 & 16 & 38 & 0.87 \\
        \midrule
        \textit{Learned Margins} & & & & & & \\
        \quad w/o structural cond. 
        & 13 & 14 & 16 & 24 & 34 & 0.79 \\
        \quad w/ structural cond.  
        & 15 & 17 & 21 & 42 & 54 & 0.85 \\
        \bottomrule
        \end{tabularx}
\end{table}

\subsection{Ordinal Diffusion Model Generates Images with Realistic Morphological Features and Disease Lesions}  
\begin{figure}[t]
    \centering
    \includegraphics[width=\textwidth]{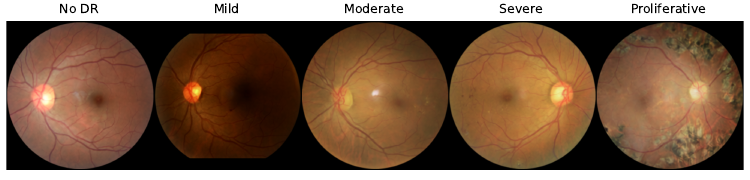}
    \caption{CFPs generated from the ordinal diffusion model with equidistant margins are realistic and capture disease related lesions.}
    \label{fig:samples}
\end{figure}

Across all classes, the model reproduced core retinal structures (Fig.~\ref{fig:samples}). The optic disc had correct brightness, shape, and placement, and blood vessels had realistic branching and tapering. Disease-specific features increased with severity: ``no DR'' samples lacked lesions; ``mild DR'' showed sparse microaneurysms; moderate and severe stages had more hemorrhages and exudates; and proliferative DR featured dense, irregular vascular formations resembling neovascularization. Some proliferative samples contained artifacts (fragmented vessels or blurriness), consistent with the higher FID values and highlighting the challenge of modeling advanced pathology. The model showed no mode collapse, with each image presenting a unique vessel and lesion distribution, demonstrating realistic fundus generation across DR stages.

\subsection{Structure Encoder Enables Counterfactual Image-to-Image Generation}
\begin{figure}[t]
    \centering
    \includegraphics[width=1.0\textwidth]{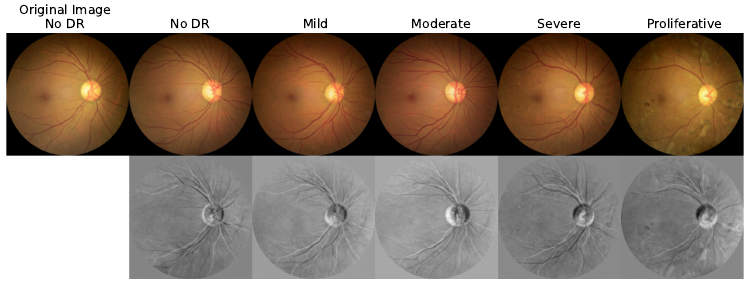}
    \caption{The structure encoder enabled counterfactual image-to-image generation. Original CFP of healthy eye on the left, generated images with ascending disease severity to the right. Below: difference image for each generated image from the healthy image.}
    \label{fig:res3}
\end{figure}

To further investigate potential benefits of adding the structural conditioning to the model, we used the ordinal diffusion model for counterfactual image generation based on an input image. Starting from a healthy reference image, we generated fundus images with ascending DR severities (Fig.~\ref{fig:res3}). The resulting images retained largely the same anatomical structures --- vasculature, optic disc, and macula --- while progressively exhibiting disease-specific lesions. Mean difference maps over the three color channels highlighted the added pathological features (Fig.~\ref{fig:res3}, second row), demonstrating that the model could modify DR characteristics while maintaining consistent structural anatomy.

\subsection{Ordinal Diffusion Model Captures Continuous Disease Structure}
\begin{figure}[t!]
    \centering
    \includegraphics[width=\textwidth]{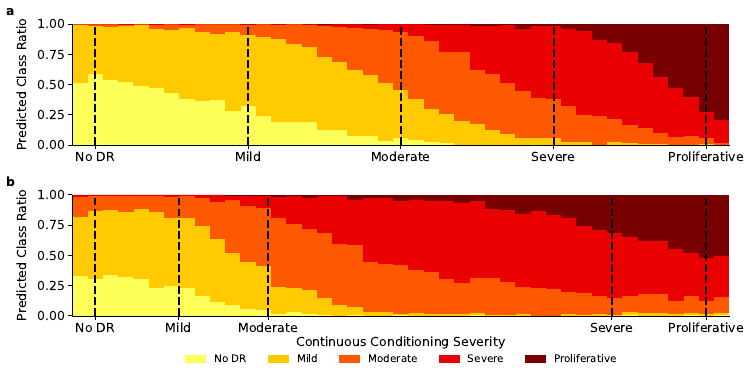}
    \caption{The ordinal diffusion model learns the continuous disease spectrum underlying ordinal labels. Images generated at intermediate class values for (\textbf{a}) the equidistant margin model and (\textbf{b}) the learned margins model were generated with a multi-class DR classifier. Resulting label proportions are shown.}
    \label{fig:res2}
\end{figure}

To examine how the model captured the continuous disease spectrum underlying the discrete DR stages, we interpolated across the ordinal spectrum and generated class-conditioned images on intermediate numbers between the disease stages (Fig.~\ref{fig:res2}). Importantly, this interpolation is not possible with the categorical baseline, as one-hot encoding does not define a continuous conditioning space. For the equidistant-margin model, these values lay between adjacent discrete numbers encoding the disease stages and for the learned-margin model, they lay between the corresponding learned values used for class conditioning. We generated 256 images for each number and classified them using our pretrained DR classifier.
We found that the model trained with equal margins showed a smooth progression between classes: as we interpolated from ``No DR'' toward ``Mild'', proportion of images classified as healthy decreased while the ``Mild'' ratio increases, and this gradual transition continued across all stages. Intermediate images contained mixed features rather than sharp jumps, showing that the model learned a continuous spectrum of the disease rather than discrete categories. For the learned-margin model, interpolation revealed smaller gaps between early stages (``No DR'' to ``Moderate'') and larger gaps for advanced stages (``Severe'' to ``Proliferative''), while still maintaining smooth transitions. The spacing reflects the model's perception of disease severity: early transitions appeared more subtle, whereas later stages corresponded to more pronounced pathological changes such as neovascularization. Thus, while equidistant margins improved quantitative performance, learned margins provide additional insights into the clinical complexity of DR progression.

\section{Discussion}
Our model generates realistic interpolations between adjacent disease stages, indicating that it learns a continuous representation of disease severity consistent with the progressive nature of DR.
Ordinal conditioning thus improves both disease-consistent generation and visual realism. Despite these strengths, several limitations remain. Learned margins make no assumptions about inter-stage distances, allowing the model to learn inter-stage geometry from data, but may suffer from training instability due to the additional learnable parameters and insufficient gradient signal from minority classes. While performing on par with the baseline, learned margins provided improved interpretability through clinically meaningful margins. Beyond architectural choices, class imbalance poses a persistent challenge: higher FID values were observed for minority classes, which were likely primarily driven by class imbalance, as the model receives substantially less gradient signal from these stages. While the model reproduces major vessel structures, finer vessel branching remains imperfect, particularly in advanced stages where neovascularization introduces complex topology. These generation artifacts point to a broader methodological challenge: evaluating medical image generators remains difficult. FID measures realism but does not guarantee clinical correctness~\cite{wu2025fid}. To address this, we evaluated a DR classifier on the generated images to assess disease consistency. Interestingly, the DR classifier performed better on generated images than on real data, suggesting that the generator may emphasize dominant class features or that the EyePACS dataset contains substantial label noise, as reported previously~\cite{juImprovingMedicalImage2021}. Furthermore, our evaluation mainly relied on automated metrics, in contrast to other recent work on generative diffusion models for CFPs~\cite{ilanchezian2025development}. Addressing these limitations motivates several directions for future work. Since ordinal conditioning is resolution-agnostic, higher-resolution synthesis is a natural next step to better capture fine pathological details. Future work should further include expert clinical assessment, quantitative evaluation of counterfactual images, and latent space analysis of the structural encoder. Finally, extending the framework to longitudinal datasets may enable explicit modeling of temporal disease progression.

\begin{credits}
\subsubsection{\ackname} This project was supported by the Hertie Foundation and by the Deutsche Forschungsgemeinschaft under Germany's Excellence Strategy with the Excellence Cluster 2064 ``Machine Learning — New Perspectives for Science'', project number 390727645. PB is a member of the Else Kröner Medical Scientist Kolleg ``ClinbrAIn: Artificial Intelligence for Clinical Brain Research''. The authors thank the International Max Planck Research School for Intelligent Systems (IMPRS-IS) for supporting SM.
\subsubsection{\discintname} The authors have no competing interests to declare that are relevant to the content of this article.
\end{credits}

\bibliographystyle{splncs04}
\bibliography{references.bib}

\end{document}